\documentclass[10pt,twocolumn,letterpaper]{article}

\usepackage{cvpr}
\usepackage{times}
\usepackage{epsfig}
\usepackage{graphicx}
\usepackage{amsmath}
\usepackage{amssymb}

\usepackage{subcaption}
\usepackage{makecell}
\usepackage{stfloats}
\usepackage{multirow}
\usepackage{bm}
\usepackage{mathrsfs}
\usepackage{arydshln}
\usepackage{floatrow}
\usepackage{booktabs}

\newfloatcommand{capbtabbox}{table}[][\FBwidth]
\newcommand{\tabincell}[2]{\begin{tabular}{@{}#1@{}}#2\end{tabular}}

\usepackage[breaklinks=true,bookmarks=false]{hyperref}

\cvprfinalcopy 

\def\cvprPaperID{****} 

\begin{document}

\title{Progressive Learning of  Low-Precision  Networks}

\author{Zhengguang Zhou$^{1}$\thanks{This work is done when Zhengguang Zhou is an intern in Intellifusion Inc.}, Wengang Zhou$^{1}$, Xutao Lv$^{2}$, Xuan Huang$^{2}$, Xiaoyu Wang$^{2}$, Houqiang Li$^{1}$\\
$^{1}$CAS Key Laboratory of GIPAS, University of Science and Technology of China\\
$^{2}$Intellifusion Inc.\\
{\tt\footnotesize zhgzh164@mail.ustc.edu.cn, \{zhwg, lihq\}@ustc.edu.cn, \{lvxutao, huang.xuan.intellif, fanghuaxue\}@gmail.com }
\and
}

\maketitle

\begin{abstract}
  Recent years have witnessed the great advance of deep learning in a variety of vision tasks. Many state-of-the-art deep neural networks suffer from large size and high complexity, which makes it difficult to deploy  in resource-limited platforms such as mobile devices.
  To this end, low-precision neural networks are widely studied which quantize weights or activations into the low-bit format.
  Though being efficient,  low-precision networks are usually  hard to train and encounter severe accuracy degradation.
  In this paper, we propose a new training strategy through expanding low-precision networks during training and removing the expanded parts for network inference.
  First, we equip each low-precision convolutional layer with an ancillary  full-precision convolutional layer based on a low-precision network structure, which could guide the network to  good local minima.
  Second, a decay method is introduced to reduce the output of the added full-precision convolution gradually, which keeps the resulted topology structure the same to the original low-precision one.
  Experiments on SVHN, CIFAR and ILSVRC-2012 datasets prove that the proposed method can bring faster convergence and higher accuracy for low-precision neural networks.

\end{abstract}

\section{Introduction}

Deep neural networks (DNNs) have achieved great progress in a variety of computer vision tasks \cite{he2016deep, Krizhevsky2012ImageNet, long2015fully,  ren2015faster,  rene2017temporal, xiao2016learning, zhu2016face}. The remarkable performance of DNNs usually requires powerful hardware with large memory and computing resources. On the other hand, there are growing demands to bring artificial intelligence to different platforms, such as  mobile phones, smart glasses, and automatic driving cars. However, most embedded devices have limited computational and memory resources  that make DNNs  difficult to deploy. Fortunately, there is a significant parameter redundancy in DNNs \cite{Shakibi2013Predicting} and a variety of model compression and acceleration methods are developed, such as network pruning \cite{han2015learning, he2017channel, li2016pruning, liu2017learning, mallya2017packnet, yu2017nisp}, network quantization \cite{Courbariaux2015BinaryConnect, faraone2018syq, Li2016Ternary,  park2017weighted, Rastegari2016XNOR, tang2017train, tung2018clip, wang2018two, Zhou2016DoReFa}, model distillation \cite{hinton2015distilling, romero2014fitnets}, low-rank approximation \cite{jaderberg2014speeding, lebedev2014speeding, zhang2015efficient}, compact network design \cite{howard2017mobilenets, iandola2016squeezenet,sandler2018mobilenetv2,  xiangyu2016shufflenet}, \emph{etc}.

\begin{figure}[t]
\centering
\includegraphics[scale =0.82]{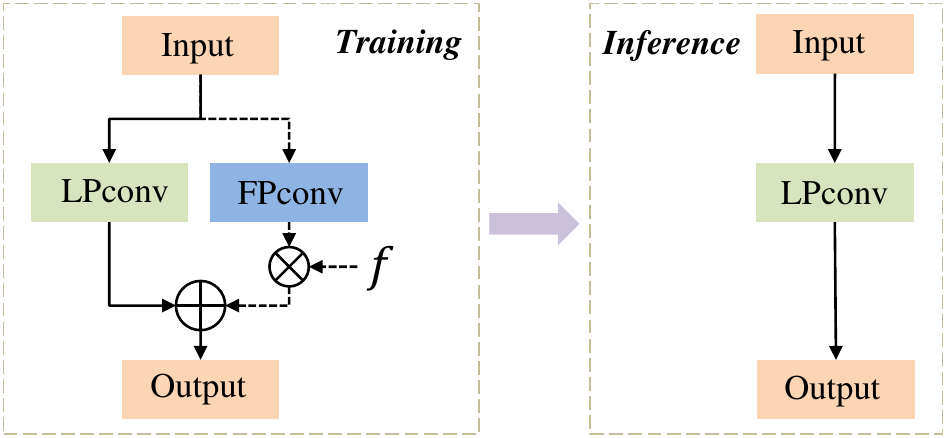}
\caption{An illustration of the proposed method. Based on a low-precision network, we equip each low-precision convolutional layer  with another full-precision one  during training. A decreasing factor $f$ is used to reduce the output of full-precision layer gradually to zero. The full-precision part is removed for network inference finally.}
\label{help_growth}
\end{figure}

As the   expensive floating point multiply-accumulate operations can be done by low-bit operations, fixed-point quantization is an effective approach to reduce both model size and computation complexity. By turning both weights and internal activations into binary,  the BNN-Net \cite{Courbariaux2016Binarized} and XNOR-Net \cite{Rastegari2016XNOR} are proposed. Hence the convolutions can be replaced with $xnor$ and $bitcount$ operations, and the network achieves $\scriptsize{\sim}32\times$ compression and $\scriptsize{\sim}58\times$ speed up in CPUs. However, the  low-precision networks are usually difficult to train and encounter severe accuracy degradation. For example, BNN-Net  and XNOR-Net drop the top-1 accuracy for AlexNet \cite{Krizhevsky2012ImageNet} by 28.7\% and 12.4\%, respectively.

The main problem is that the standard training algorithm is not effective due to the low representational capability of  low-precision networks.
Assume the weights or activations are quantized to $k$-bit ($k$ usually smaller than $8$), all weights or activations in a layer could only choose values in a finite set whose size is just $2^k$. Meanwhile, the $32$-bit full-precision  weights or activations can choose any value in a set of size $2^{32}$ which is far greater than $2^k$. Thus, the feature maps of low-precision networks cannot be well represented with limited values.
On the other hand, the parameters of low-precision networks are not well updated during training due to the non-differentiable  quantization functions. Therefore,  the low-precision networks could fall into bad local minima.

To deal with this problem,  we propose a general framework to progressively guide the learning of a low-precision network. As shown in Figure \ref{help_growth} (left), we equip each low-precision convolutional layer with an additional full-precision convolutional layer, and the output of the expanded layer is the combination of two different precision outputs. In this way, the representational capability of the output in the expanded layer is significantly enhanced which could enable the network with faster convergence and  better local minima.  In addition, the  full-precision layers are not expected for inference. Therefore, we propose a  decay method to gradually weaken the output of the full-precision convolution, \emph{i.e.,} we use a monotonically decreasing factor ($f$) related to the training step to scale each output. After training, the  factor is weakened to zero and all full-precision convolutional layers can be safely removed resulting in the original low-precision structure (see Figure \ref{help_growth}).
As a result, we call the expanded network \textbf{EXP-Net}.

Overall, the contributions of this paper are two-fold:
1) We present a general framework to guide the learning of low-precision networks and the resulted structure is kept the same as the original one with our decay method.
2) Extensive experiments on SVHN, CIFAR and ILSVRC-2012 datasets demonstrate the effectiveness and competitiveness of our method.

\section{Related Work}
Network compression and acceleration methods have been widely developed by the community. In this section, we mainly review three related categories of methods, \emph{i.e.,} network pruning, network quantization and network distillation.

\textbf{Network pruning}.
With the observation that DNNs have a significant parameter redundancy, pruning methods have been widely studied
for reducing network complexity.
On the one hand, the  parameters can be pruned on a pre-trained network.
Han \emph{et al.}   prune weights without significant loss in accuracy \cite{han2015learning}, but this method requires specific hardware for acceleration. Thus, many other works focus on structured pruning which results in the structured sparse network. In \cite{li2016pruning}, Li \emph{et al.} prune the unimportant filters whose absolute weights are small. He \emph{et al.}  prune channels based on LASSO regression and reconstruct output feature maps with remaining channels for a pre-trained CNN effectively \cite{he2017channel}.
On the other hand, many methods implement pruning by adding regularization or other modifications during training. Liu \emph{et al.} impose channel sparsity by imposing $\ell_1$ regularization on the scaling factors in batch normalization.
In \cite{he2018soft}, He \emph{et al.} propose a soft filter pruning method which allows the pruned filters to be updated during the training procedure.

Our proposed method can be regarded as a special kind of pruning. We equip each low-precision  convolutional layer with another  convolutional layer in parallel and transform the original network to a complex counterpart which is more powerful to learn representations. During training, the full-precision parts of the large network are gradually  weakened and expected to be pruned to recover the primary model. Thanks to the powerful  representations learning of the large network, the resulting  model can achieve a better performance compared to the one which is not expanded.

\textbf{Network quantization}.
Recently, it has shown that full-precision floating point is not necessary and 16-bit fixed point representation is enough to train the networks \cite{Gupta2015Deep}. Based on quantizing the weights only, many effective methods are proposed.
The BinaryConnect method \cite{Courbariaux2015BinaryConnect} quantizes all weights into +1 and -1 using sign function and exhibits comparable performance on small dataset. Binary Weight Network (BWN \cite{Rastegari2016XNOR}) introduces several scaling factors on the binary filters in each layer  and  achieves good results on a large  dataset. In another way, the Incremental Network Quantization (INQ \cite{Zhou2016Incremental}) method gradually turns all weights into a logarithmic format in an incremental manner.

Further, by turning both weights and  activations into the low-bit format, the network achieves much compression and speedup.
The DoReFa-Net \cite{Zhou2016DoReFa} investigates the effect of different bitwidths for quantizing weights, activations as well as gradients and achieves comparable accuracy compared with full-precision networks. Another fine-grained quantization approach compared to DoReFa-Net is proposed in  \cite{faraone2018syq} (called SYQ), which partitions the quantization into weight subgroups and  introduces pixel-wise or row-wise scaling on them.
As SYQ introduces more parameters and improves the model capacity compared to DoReFa-Net, it enables the low-precision network to find a better sub-optimal solution. In another way, the Bi-Real Net \cite{liu2018bi} connects the real activations  (after batch normalization layer and before the sign function) to activations of the consecutive block through an identity mapping and significantly enhances the representational capability. Besides, Zhuang \emph{et al.} propose three effective approaches to improve low-precision network training  \cite{zhuang2018towards}.

Unlike the above quantized methods which develop a new quantization function to get a better quantized model, our proposed method is a better training strategy  to progressively train a low-precision network,  which is beyond quantization function. This is achieved by improving the representational capability and diversity using a mix of low-precision and full-precision structure during training. The resulted structure is kept the same as the original one rather than Bi-Real Net \cite{liu2018bi}.

\textbf{Network distillation}. Network distillation first trains a big network (\emph{i.e.,} teacher network)  and then trains a shallow one (\emph{i.e.,} student network) to mimic the output distributions of the teacher. Hinton \emph{et al.} propose knowledge distillation to soften the output class probabilities which contain more information than the one-hot label \cite{hinton2015distilling}. In \cite{romero2014fitnets}, Romero \emph{et al.} incorporate the intermediate feature maps of the teacher model as hints to train a deeper and thinner student network.
 Rather than  transferring a pre-trained teacher to a student, Zhou \emph{et al.} propose a unified framework that exploits a booster net to help train the lightweight net for inference \cite{Zhou2017Rocket}. Moreover, Zhang \emph{et al.} propose  deep mutual learning (DML) strategy, where two networks learn collaboratively and teach each other \cite{Zhang2017Deep}.
Using the knowledge distillation technique, Apprentice \cite{mishra2017apprentice} improves low-precision network accuracy significantly.

 Our proposed method can also be regarded as an implicit teacher-student learning, which is different from the works in \cite{mishra2017apprentice,zhuang2018towards}. The added full-precision part is the teacher network and the original low-precision part is the student network. During training, the powerful teacher part guides the student part learning and brings the power to the low capability student part gradually.

\begin{figure}[htbp]
\centering
\includegraphics[scale =0.84]{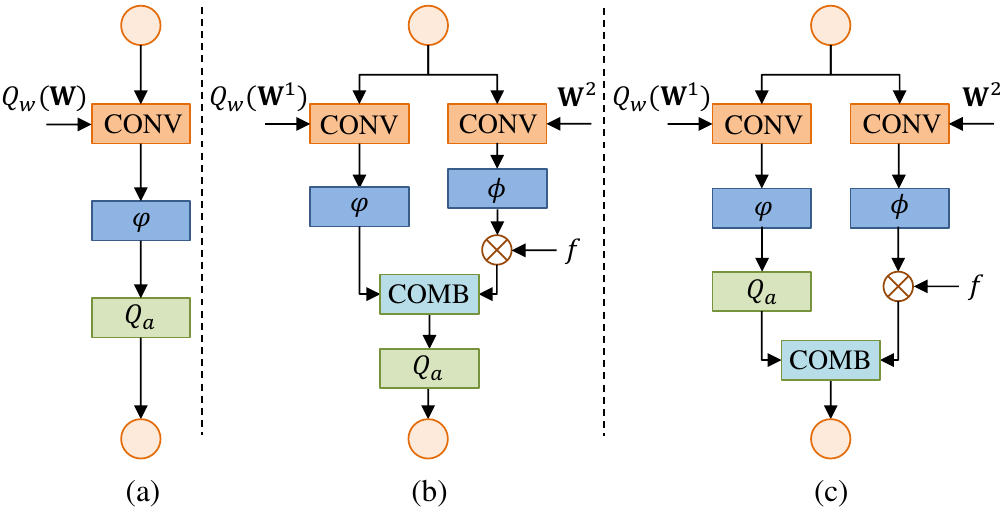}

\caption{The building blocks of  an original low-precision network (a) and our two  schemes (b and c) for expanding network.  $Q_w$ and $Q_a$ are the quantization functions for weights and activations, respectively. ``COMB" denotes the combined operation of two outputs including addition, subtraction and concatenation.  The quantization function could be taken from DoReFa-Net \protect\cite{Zhou2016DoReFa}, XNOR-Net \protect\cite{Rastegari2016XNOR}  or SYQ \protect\cite{faraone2018syq}. $\varphi$ and $\phi$ are two nonlinear functions. $f$  is a monotonically decreasing factor to reduce the outputs of the full-precision parts gradually.}
\label{scheme_abc}
\end{figure}

\section{Method}
In this section we first introduce the fundamental of low-precision neural networks. Then, we analyze the representational capability of low-precision networks. After that, we elaborate the  formulation of our proposed method which implicitly guides the learning of low-precision networks.

\subsection{Review of Standard Low-Precision Networks}

Considering a typical deep neural network with $L$ layers, let$\mathbf{W}_l$  and $\hat{\mathbf{A}}_{l-1}$ denote the convolutional kernel  and input in the $l$th layer. The building block of this layer takes $\mathbf{W}_l$  and $\hat{\mathbf{A}}_{l-1}$ as the inputs and gets the output $\hat{\mathbf{A}}_l$ which is also the input of the next layer. The main procedure to train the low-precision network in the $l$th layer  is given by Eq. (\ref{scheme_a}).
\begin{small}
\begin{equation}\label{scheme_a}
\left\{
\begin{array}{lr}
\hat{\mathbf{W}}_l = Q_w(\mathbf{W}_l),  & \\
\mathbf{Z}_l = \hat{\mathbf{W}}_l \otimes \hat{\mathbf{A}}_{l-1}, & \\
\mathbf{A}_l = \varphi (\mathbf{Z}_l), & \\
\hat{\mathbf{A}}_l = Q_a (\mathbf{A}_l),  &
\end{array}
\right.
\end{equation}
\end{small}where $Q_w$ and $Q_a$ are two quantization functions to turn weights $\mathbf{W}_l$ and activations $\mathbf{A}_l$ into low-bit format, respectively.
$\varphi$ is the nonlinear function to restrict the range of activations and $\otimes$ is the convolutional operation.
Figure \ref{scheme_abc} (a) exhibits the above processes clearly.

The quantization function plays an important role in low-precision networks and a good function can achieve significant accuracy improvement \cite{cai2017deep, faraone2018syq, park2017weighted}.
For example, BNN-Net \cite{Courbariaux2016Binarized} quantizes weights to $-1$ and $1$ using the sign function while XNOR-Net \cite{Rastegari2016XNOR} introduces a filter-wise scaling factor  for each binary filter, which can expand the capacity of binary neural networks.
Since our work is to present a general framework to progressively train low-precision networks, our method is beyond designing quantization functions (\emph{e.g.,} DoReFa-Net \cite{Zhou2016DoReFa}, XNOR-Net \cite{Rastegari2016XNOR}  and SYQ \cite{faraone2018syq}). Besides, our approach is also independent of and complementary with other training techniques (\emph{e.g.,} \cite{wang2018two,zhuang2018towards}).

\subsection{The Representational Capability of  Low-Precision Networks}
Here, we analyze the representational capability of a standard low-precision neural network in one layer.
We denote $D(\mathbf{z})$ as the number of all possible different values of $\mathbf{z}$ and the representational capability $\mathbb{R}(\mathbf{z})$ as the number of all possible configurations of $\mathbf{z}$, where $\mathbf{z}$ could be a tensor with any dimension. Thus,  $\mathbb{R}(\mathbf{z})=(D(\mathbf{z}))^s$ where $s$ is the size of $\mathbf{z}$ \cite{liu2018bi}.
For simplicity, we constrain the bitwidth to be 1 (\emph{i.e.,} the 1-bit network) and let the size of weights be $d\times d\times M\times N$ and activations be $H\times W\times M$.
Therefore, the quantized weights $\hat{\mathbf{W}}_l$ and the inputs $\hat{\mathbf{A}}_{l-1}$ of the $l$th layer have both only two different values among $\left\{-1,+1 \right\}$, \emph{i.e.,} $D(\hat{\mathbf{W}}_l)=D(\hat{\mathbf{A}}_{l-1})=2$. And the representational capability  $\mathbb{R}(\hat{\mathbf{A}}_{l-1})$ is $2^{H\times W\times M}$.

Then we calculate the number of unique values in the quantized convolutional output (\emph{i.e.,} $D(\mathbf{Z}_l)$). The $i$th value in $\mathbf{Z}_l$ could be computed by the dot product of two binary vectors $\mathbf{x}$ and $\mathbf{y}$ as $z_l^i=\mathbf{x} \cdot \mathbf{y}$,
where $\mathbf{x}$ is vectorization of a $d\times d\times M$ patch of $\hat{\mathbf{A}}_{l-1}$ and $\mathbf{y}$ is also vectorization of a corresponding filter of size  $d\times d\times M$. Since $\mathbf{x}$ and $\mathbf{y}$ are vectors of $\left\{-1,+1\right\}$, the dot product can be replaced by $xnor$ and $bitcount$ operations \cite{Zhou2016DoReFa}:
\begin{small}
\begin{equation}\label{xnor}
\mathbf{x} \cdot \mathbf{y}=d\times d\times M-2\times bitcount(xnor(\mathbf{x},\mathbf{y})),
\end{equation}
\end{small}where $bitcount$ counts the number of ones in a vector and $xnor$ is the logic operation. Since both $\mathbf{x}$ and $\mathbf{y}$ have many different configurations, $bitcount(xnor(\mathbf{x},\mathbf{y}))$ can be any integer from $0$ to $d\times d\times M$ possibly. Thus, the number of different values in $\mathbf{Z}_l$ is the same as the number of integers in $[0, d\times d\times M]$, which is $D(\mathbf{Z}_l)=d\times d\times M+1$. Moreover, the nonlinear function and batch normalization layer do not change the number of unique values and representational capability of $\mathbf{Z}_l$. Table \ref{repre} shows the number of all possible different values $D(\mathbf{z})$ after each operation in one layer.

\begin{table}
	\small
	\begin{center}
		\begin{tabular}{|c|c|c|c|c|c|c|}
			\hline
			\multicolumn{1}{|c|}{\multirow{2}{*}{Operation}} &
			\multicolumn{1}{c|}{\multirow{2}{*}{FP-Net}} &
			\multicolumn{1}{c|}{\multirow{2}{*}{LP-Net}} & \multicolumn{2}{c|}{\multirow{1}{*}{Scheme-1}} & \multicolumn{2}{c|}{\multirow{1}{*}{Scheme-2}} \\
			\cdashline{4-7}[0.8pt/2pt]
			& & & LP & FP & LP & FP \\
			\hline 
			Input & $|\mathbb{Q}|$ & 2 &  \multicolumn{2}{c|}{2} &  \multicolumn{2}{c|}{$|\mathbb{Q}|$} \\
			\cdashline{1-7}[0.8pt/2pt]
			Conv & $|\mathbb{Q}|$  & 289 & 289 & $|\mathbb{Q}|$  &  $|\mathbb{Q}|$ &  $|\mathbb{Q}|$\\
			\cdashline{1-7}[0.8pt/2pt]
			BN/$\phi$ & $|\mathbb{Q}|$ & 289 & 289 & $|\mathbb{Q}|$  &  $|\mathbb{Q}|$ &  $|\mathbb{Q}|$\\
			\cdashline{1-7}[0.8pt/2pt]
			Add & - & - &  \multicolumn{2}{c|}{$|\mathbb{Q}|$}  &  \multicolumn{2}{c|}{-}  \\
			\cdashline{1-7}[0.8pt/2pt]
			$Q_a$ & $|\mathbb{Q}|$ & 2 & \multicolumn{2}{c|}{2} & 2 & - \\
			\cdashline{1-7}[0.8pt/2pt]
			Add & - & - & \multicolumn{2}{c|}{-} & \multicolumn{2}{c|}{$|\mathbb{Q}|$}  \\
			\hline
		\end{tabular}
	\end{center}
\vspace{-0.2cm}
\caption{The  number of all possible different values ($D(\mathbf{z})$) of the full-precision network, its low-precision counterpart and our two expanding schemes during training. $\mathbb{Q}$ indicates the set from which each  full-precision (\emph{i.e.,} $32$-bit) element of $\mathbf{z}$ can choose, \emph{i.e.,} the size of the set $|\mathbb{Q}| = 2^{32}$. Here, we constrain the bitwidth to 1 and let the size of weights $\mathbf{W}_l$ be $3\times 3\times 32\times 64$. So $D(\mathbf{Z})$ in low-precision network is $3\times 3\times 32+1=289$. ``LP" and ``FP" denote the low-precision and full-precision parts, respectively. ``-" indicates the operation is absent in the method.  }\label{repre}
\end{table}

\begin{table}
\small
\begin{center}
\begin{tabular}{|c|ccc|}
\hline
 Model size & Weights & Feature maps &  $\mathbb{R}(\mathbf{Z})$      \\
 \hline
  Large ($1\times$) & $3\times 3\times 32\times 64$  &  $ 8\times 8\times  64$  &  $289^{4096}$  \\
  Small ($0.5 \times$) & $3\times 3\times 16\times 32$  &  $ 8\times 8\times  32$  &  $145^{2048}$  \\
  \hline
\end{tabular}
\end{center}
\vspace{-0.2cm}
\caption{ The representational capability $\mathbb{R}(\mathbf{Z})$ related to different size of weights and feature maps in different models. The small model is derived from the large one by shrinking the number of filters for every layer by 50\%.  Here, the weights and activations are quantized to 1-bit. }\label{size_rp}
\end{table}
Model size plays an important role in low-precision networks. As shown in Table \ref{size_rp}, the representational capability $\mathbb{R}(\mathbf{Z})$ in the large model is more powerful than that in the small one, the former has enough power to learn representations even for 1 bitwidth. On the contrary, smaller models encounter severe accuracy degradation due to their limited representational capability. Consequently, our method is designed for compact models rather than large networks.

\subsection{EXP-Net}
From Table \ref{repre}, we can see the representational capability of a low-precision network is much lower than the full-precision one, which causes the network  hard to train and large accuracy degradation.
Therefore, we propose EXP-Net to implicitly guide training the low-precision network by expanding the network and improving the representational capability.
This is achieved by equipping each low-precision  convolutional layer with another full-precision convolutional layer in parallel, and the output of the expanded layer is the combination of two different precision outputs, which is shown in Figure \ref{help_growth} (left). Considering the order between activation quantization and the element-wise add operation, we have  two different expanding schemes to construct our network structure. And the decay method used in the expanding schemes is given at last.

\textbf{Scheme-1}. Figure \ref{scheme_abc} (b) shows our scheme-1, where we quantize the activations after combining the low-precision and full-precision parts. The forward path can be formulated as Eq. (\ref{scheme_b}).
\begin{small}
\begin{equation}\label{scheme_b}
\left\{
\begin{array}{lr}
\mathbf{Z}_l^1 = Q_w(\mathbf{W}_l^1) \otimes \hat{\mathbf{A}}_{l-1},  &\\
\mathbf{Z}_l^2 = \mathbf{W}_l^2 \otimes \hat{\mathbf{A}}_{l-1},   &\\
\mathbf{A}_l = Combine[\varphi (\mathbf{Z}_l^1), \phi (\mathbf{Z}_l^2) \times f],  &\\
\hat{\mathbf{A}}_l = Q_a (\mathbf{A}_l), &
\end{array}
\right.
\end{equation}
\end{small}where $\varphi$ and $\phi$ are two nonlinear functions and $f$ is the monotonically decreasing factor. ``$Combine$" could be addition, subtraction and concatenation.  
Since the weights $\mathbf{W}_l^2$ in the full-precision convolutional layer are not quantized, the output feature maps $\mathbf{Z}_l^2$ could have any real value in the set of $32$-bit full-precision number $\mathbb{Q}$ (see Table \ref{repre}). The representational capability of internal activations $\mathbf{A}_l $ after adding the two outputs  is improved significantly. This powerful representation could implicitly guide the network finding the more satisfactory  parameters compared with training the low-precision network from scratch.

\textbf{Scheme-2}. Besides scheme-1, we have another alternative scheme which is achieved by  adding the two outputs after quantizing the activations of low-precision convolution. Our scheme-2 is shown in Figure \ref{scheme_abc} (c). The main computation in the $l$th layer is formulated as Eq. (\ref{scheme_c}).
\begin{small}
\begin{equation}\label{scheme_c}
\left\{
\begin{array}{lr}
\mathbf{Z}_l^1 = Q_w(\mathbf{W}_l^1) \otimes \hat{\mathbf{A}}_{l-1}, &\\
\mathbf{Z}_l^2 = \mathbf{W}_l^2 \otimes \hat{\mathbf{A}}_{l-1}, &\\
\hat{\mathbf{A}}_l^1 = Q_a (\varphi (\mathbf{Z}^1_l)),  &\\
\hat{\mathbf{A}}_l = Combine[\hat{\mathbf{A}}_l^1, \phi (\mathbf{Z}_l^2) \times f]. &
\end{array}
\right.
\end{equation}
\end{small}Since the output tensor $\hat{\mathbf{A}}_l$ consists of $\mathbf{Z}_l^2$ which could have any real value in  $\mathbb{Q}$,  both input and output can choose any value from $\mathbb{Q}$ (see Table \ref{repre}). Therefore, the network almost becomes a full-precision counterpart. As a result, the network could converge very quickly at the beginning. But while the output of the full-precision convolution is gradually weakened to zero, the network loses much information and the convergence is broken. Fortunately, the low-precision parts have already learned reliable parameters at that moment. Thus, the model can also converge  after removing all full-precision convolutional layers and has a better performance compared with the original low-precision network trained from scratch.

As for the backward path in our schemes, we use  Straight Through Estimator (STE) \cite{bengio2013estimating} method and the main computation is the same as that in other quantization methods, such as DoReFa-Net\cite{Zhou2016DoReFa}, SYQ\cite{faraone2018syq}. Since the decreasing factor $f$  scales the outputs of all full-precision convolutions, we also need to consider it in backpropagation.

\textbf{Decay Method}. Our method aims to implicitly guide training a low-precision network by expanding it during training  and the topology structure of the network is not demanded to change for inference. Thus, we use a decay method to gradually reduce the output of the added full-precision convolution, \emph{i.e.,} a monotonically decreasing factor related to the training step is used to scale each output. The factor $f$ is required to be in $[0, 1]$ and  be zero at the end of the training.
We construct two different decay functions, \emph{i.e.,} a cosine decay function for the continuous case and an exponential decay function for the step-wise case.  Let $\alpha$ and $\beta$ denote the  global step and decay step, respectively.
The cosine decay and exponential decay functions are computed as:
\begin{small}
\begin{eqnarray}\label{scheme_f1}
f_{cos} &=& 0.5+0.5\times cos(\pi \times\frac{min(\alpha,\beta)}{\beta}),  \\
 f_{exp}&=& \left\{
\begin{array}{rcl}
\delta^\frac{\alpha}{\beta},     &   &  {\mbox{if} \quad    f_{exp}\geq \epsilon} \\
0,                                 &   &  {\mbox{else}}
\end{array} \right.,
\end{eqnarray}
\end{small}where $\delta$ is the decay rate and $\epsilon$ is the decay threshold hyperparameter in exponential decay function. Let $\frac{\alpha}{\beta}$ be an integer division, the exponential decay follows a step-wise function. The function curves  are shown in Figure \ref{decay_curve}.

\section{Experiment}
\subsection{Datasets and Settings}
\textbf{Datasets.} Four benchmark datasets for image classification are used in our experiments. 1) The SVHN dataset \cite{Netzer2012Reading}
is a real-world digit recognition dataset of size $32 \times 32$ and is obtained from house numbers in Google Street View images. There are 73,257 images for training, 26,032 images for testing and 531,131 less difficult examples which can be used as extra training data. The training images are resized to $40 \times 40$. 2) The CIFAR10 and CIFAR100 \cite{krizhevsky2009learning}  consist of a training set of 50,000 and a test set of 10,000 color images of size $32 \times 32$ with 10 classes and 100 classes, respectively. The training images are padded by 4 pixels and randomly flipped. 3) The ILSVRC-2012  \cite{russakovsky2015imagenet} is a large scale high resolution image classification dataset with 1,000 classes. It consists of 1.2 million training images and 50,000 validation images. The training images are randomly cropped and resized to $224\times 224$ with random flipping. We report our single-crop evaluation results using top-1 and top-5 accuracies.

\textbf{Implementation Details.}
1) For SVHN, we adopt the network proposed by \cite{Zhou2016DoReFa} which has seven convolutional layers and one fully connected layer and another three models are derived from the full model by shrinking the number of filters for every convolutional layer by 50\%, 75\% and 87.5\%, respectively. All the above networks are trained using ADAM \cite{kingma2014adam} with batch size 128 and 200 epochs. The initial learning rate is set to 0.001 and is divided by 5 at 100 epochs.
2) The VGG-16 network architecture derived from \cite{li2016pruning} is used on CIFAR, but we add batch normalization \cite{Ioffe2015Batch} after each convolutional layer. The network is trained using SGD with batch size 128 and 300 epochs.
The initial learning rate is set to 0.01 and is divided by 2 at 120, 150 and 180 epochs.
3) On ILSVRC-2012, we test the AlexNet \cite{Krizhevsky2012ImageNet}. We replace the local response normalization layer with batch normalization and discard the dropout layer,  which is the same as \cite{Zhou2016DoReFa}. We use SGD with momentum 0.9 and 100 epochs to train the model. The initial learning rate is set to 0.02 and is divided by 10 at 50, 70 and 90 epochs.
As for the  factor $f $ shown in Figure \ref{scheme_abc}, we use the learning rate decay functions in TensorFlow \cite{Abadi2016TensorFlow}. 
For low-precision networks, we do not quantize the first and last convolutional layers as XNOR-Net \cite{Rastegari2016XNOR} and DoReFa-Net \cite{Zhou2016DoReFa} do.
%
In all experiments, our EXP-Nets are trained from scratch without needing any pre-trained model.
Our source code will be released for public research.

\subsection{Ablation Study}

First of all, we analyze the influence of the modules or training ways in our proposed EXP-Net. There are seven aspects: 1) Decay function. 2) The combined operations of low-precision and full-precision outputs including addition, subtraction and concatenation. 3) Two scheme for constructing the building block of EXP-Net. 4) Model size. 5) Synchronization or asynchronization for weakening each full-precision output. 6) Other expanded structures. 7) Quantization methods used in EXP-Net.
Our default settings are cosine decay function, addition, scheme-2, synchronization.

\subsubsection{Decay Function.}

\begin{figure}
	\begin{floatrow}
		\ffigbox{%
           \includegraphics[scale =0.27]{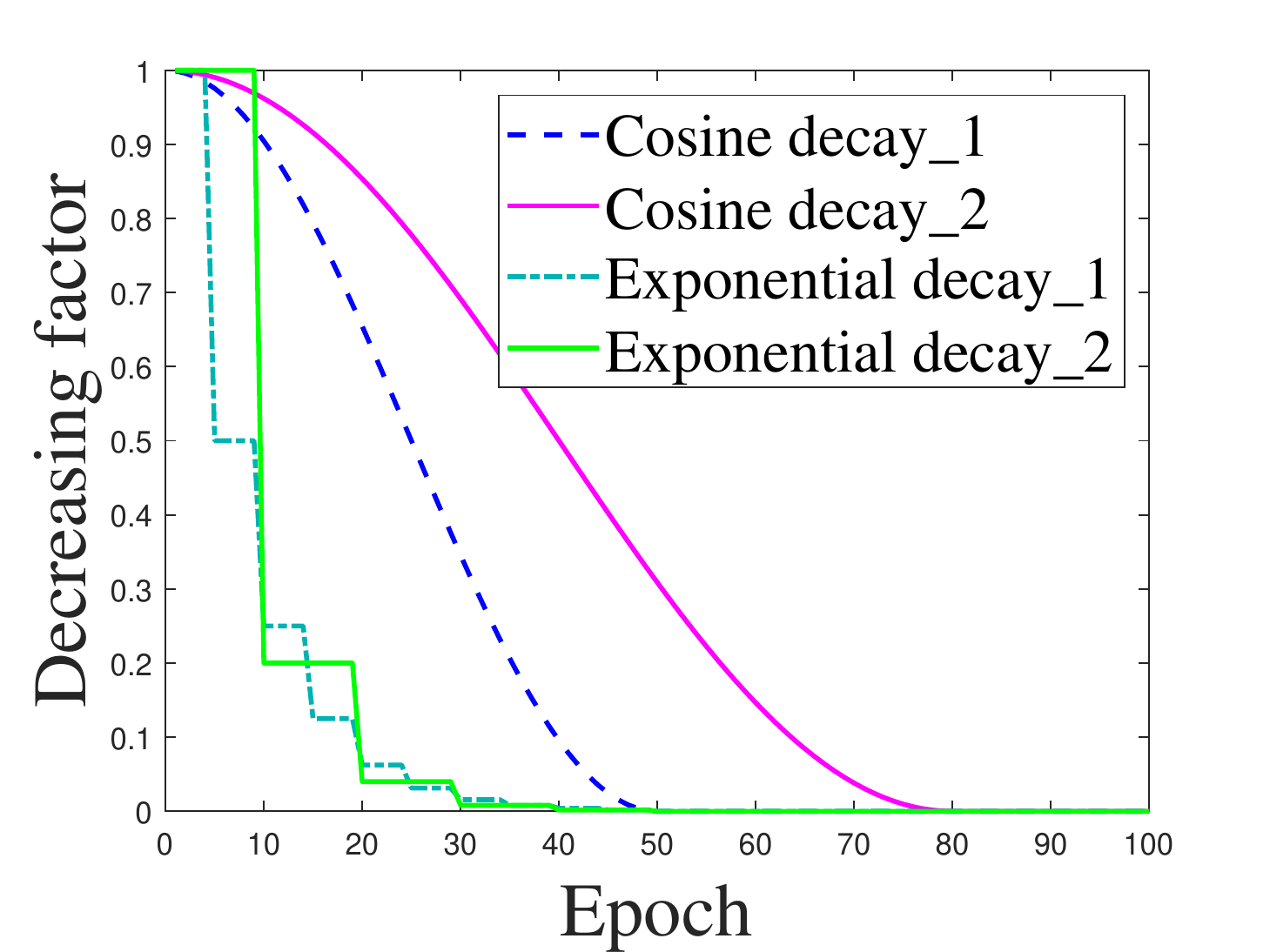}
		}{%
			\vspace{-0.2cm}
			\caption{The curves of four different decay functions used in Figure \ref{decay_vgg}.}\label{decay_curve}%
		}
		
		\capbtabbox{%
			\small
			\begin{tabular}{ccc}
				\toprule[2pt]
				Function & $\alpha$ & $\delta$ \\ \midrule[1pt]
				Cosine 1 & 50 & -  \\
				Cosine 2 & 80 & -  \\ \midrule[1pt]
				Exponential 1 & 5 & 0.5  \\
				Exponential 2 & 10 & 0.2  \\
				\bottomrule[2pt]
			\end{tabular}
		}{%
			\caption{The parameters of four decay functions used in Figure \ref{decay_vgg}.}\label{decay_parameter}%
		}
	\end{floatrow}
\end{figure}

\begin{figure}[t]
	\begin{minipage}[b]{.45\linewidth}
		\centering
\includegraphics[scale =0.3]{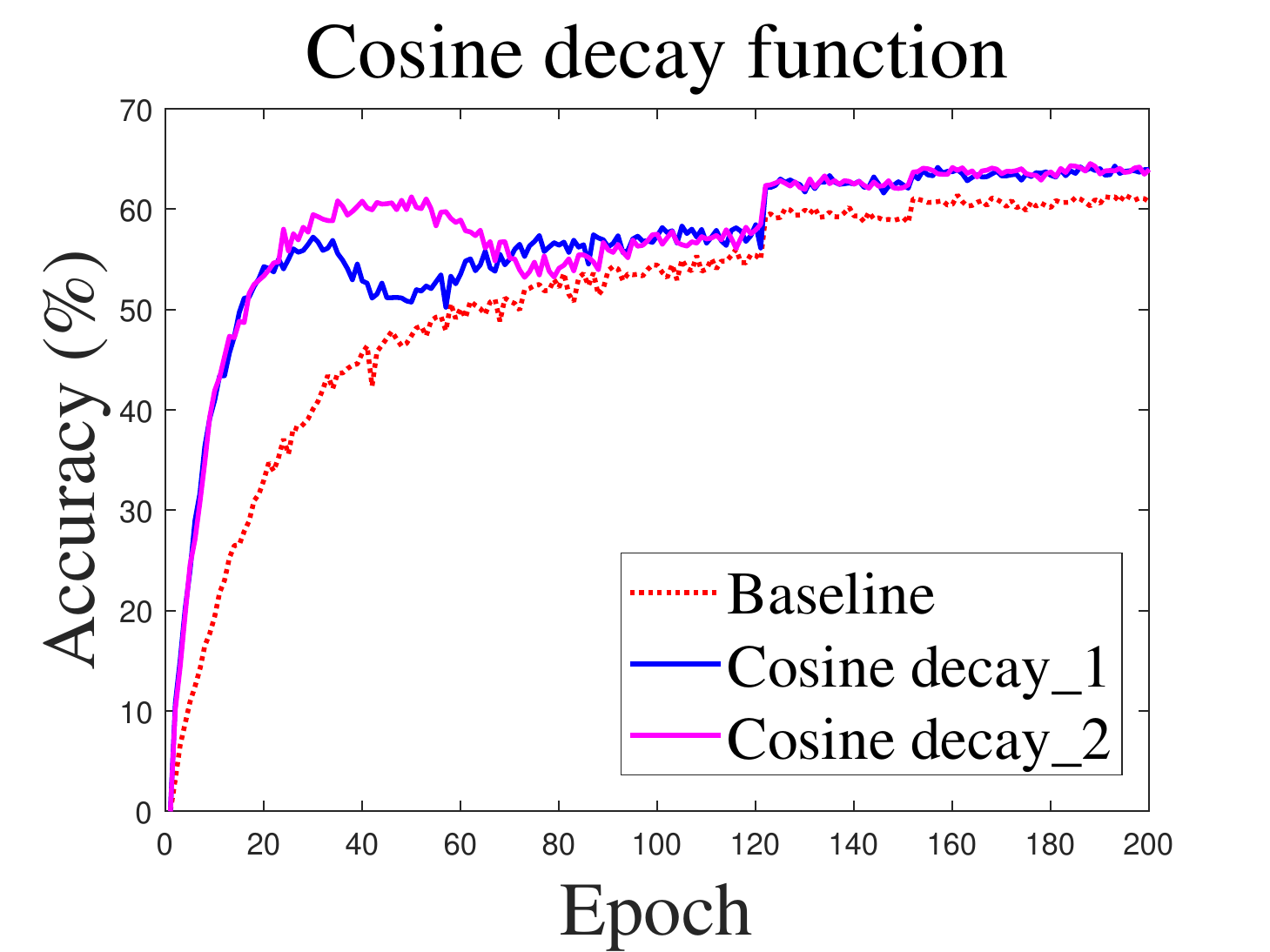}
	\end{minipage}
	\hspace{0.2cm}
	\begin{minipage}[b]{.45\linewidth}
		\centering
\includegraphics[scale =0.3]{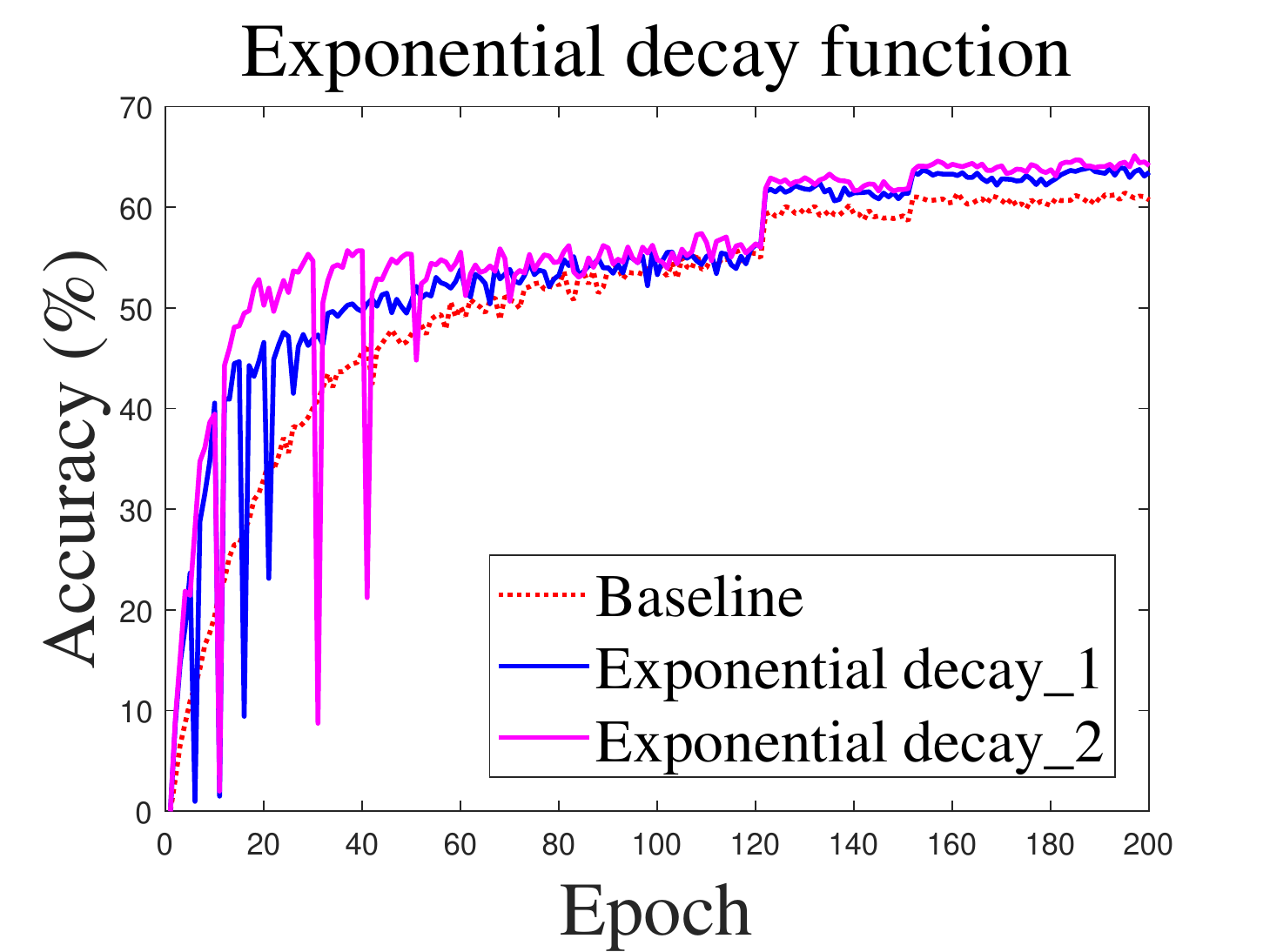}
	\end{minipage}
	\vspace{-0.2cm}
	\caption{Validation accuracies of the VGG-16 network on CIFAR100 dataset with the baseline method (DoReFa-Net \cite{Zhou2016DoReFa}) and our scheme-2. Here, both weights and activations are quantized to 1-bit. The decay functions are shown in Figure \ref{decay_curve}. Our method is consistently better under all settings. }
	\label{decay_vgg}
\end{figure}

The decay function plays an important role in the training of our EXP-Net. Here, we construct four different decay functions and evaluate our method on CIFAR100 dataset. Table \ref{decay_parameter} and Figure \ref{decay_curve} show the parameters and curves of the decay functions, respectively.
We have two major observations from Figure \ref{decay_vgg}. First, our method can speed the convergence and find better local minima no matter what the decay function we use. Specifically, our EXP-Net can improve the baseline accuracy from 61.43\% to 64.54\%.

Second, two types of decay functions have different effects on our EXP-Net training. As shown in Figure \ref{decay_vgg} (left), the effect of cosine decay function to EXP-Net training is smooth. The gap of accuracy between adjacent epoch is not obvious. However, our EXP-Net degrades the performance constantly while the scaling factor $f$ is decreased close to zero.
On the contrary, the effect of exponential decay function to training is more significant. From Figure \ref{decay_vgg} (right), it can be seen that the network performance becomes better and better when the factor is on the steps. While it jumps from one step to the next step, the accuracy is degraded heavily. Fortunately, the degradation gets more and more invisible as the gap of adjacent steps becomes smaller and smaller.
We consider that the full-precision layers are far more important than low-precision layers and control the network outputs. Therefore, when the outputs of full-precision layers are weakened by a large step in exponential decay function or close to zero in cosine decay function, our EXP-Net loses the powerful representational capability which results in a visible increase of error.

\subsubsection{Combined Operation.}

\begin{figure}[htbp]
		\centering
\includegraphics[scale =0.55]{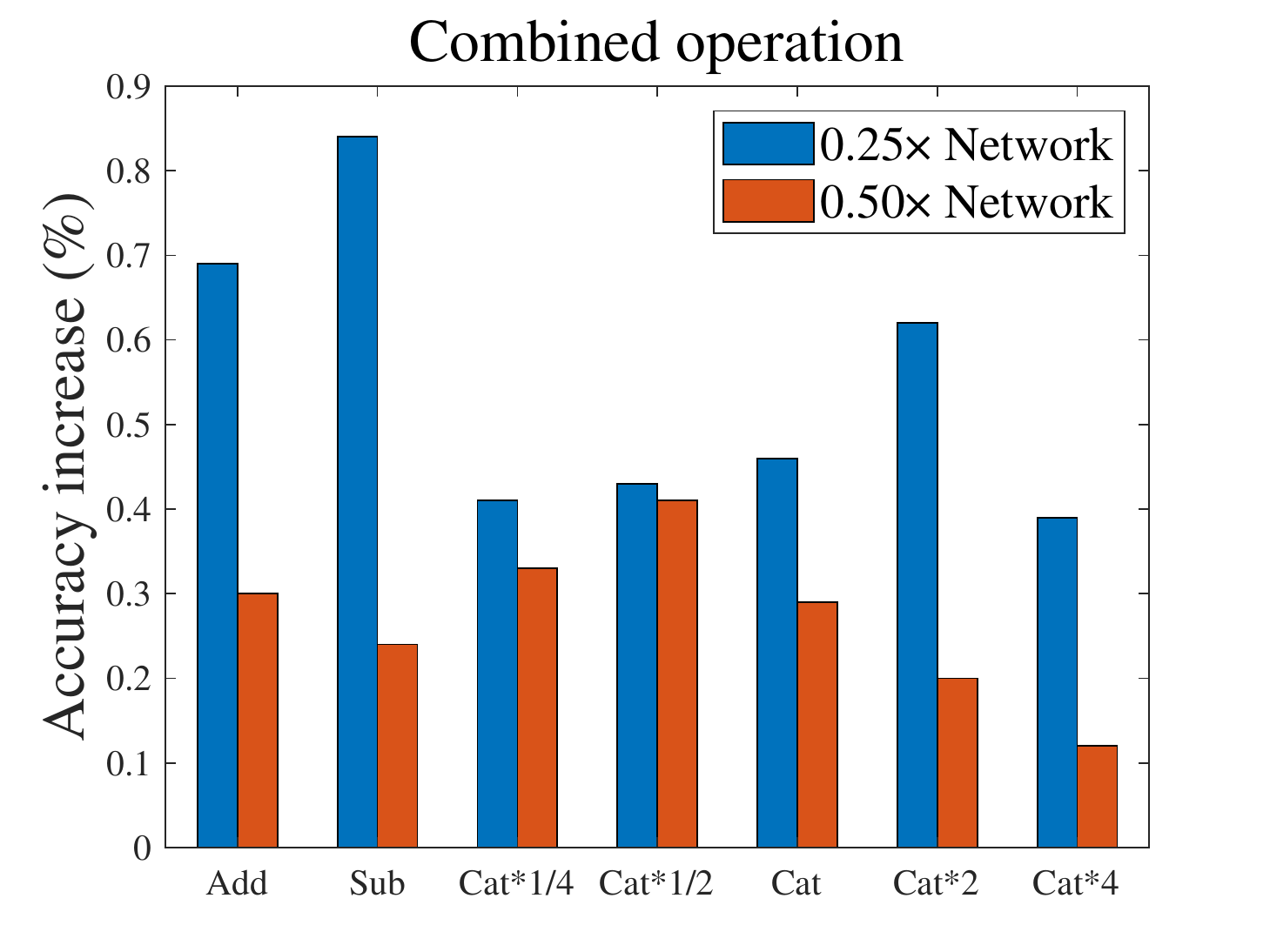}
		\vspace{-0.1in}
		\caption{Comparison accuracies of EXP-Nets using different ways for combining two outputs. The model is adopted 0.25$\times$ and 0.5$\times$ network on SVHN dataset. ``Add", ``Sub" and ``Cat" denote addition, subtraction and concatenation, respectively. ``Cat*$x$" denotes the number of filters in full-precision layers is shrunk $x$ times.}\label{add_cat}
\end{figure}

The combined operation of low-precision and full-precision convolutional outputs in EXP-Net could be addition, subtraction and concatenation. The outputs of the two layers must have the same size in addition and subtraction  while the size of the two outputs not need to be the same in the concatenation. Thus, we can vary the number of channels of the full-precision layer in the concatenation.

As shown in Figure \ref{add_cat}, all kinds of combined operations can improve the baseline accuracy. Concatenation in EXP-Net can improve the performance more than addition. Since concatenation  could increase the size of outputs while addition operation can not, the former brings more powerful representational capability than the latter. Besides, too wide or too thin full-precision layers are also not good. Too wide full-precision layer degrades the performance more when it is weakened close to zero, while too thin full-precision layer offers the limited representational capability.
Surprisedly, subtraction is also not bad. We consider it is able to enhance the representational capability like addition.

\subsubsection{Scheme-1 vs. Scheme-2.}

\begin{figure}[t]
	\begin{minipage}[b]{.45\linewidth}
		\centering
\includegraphics[scale =0.3]{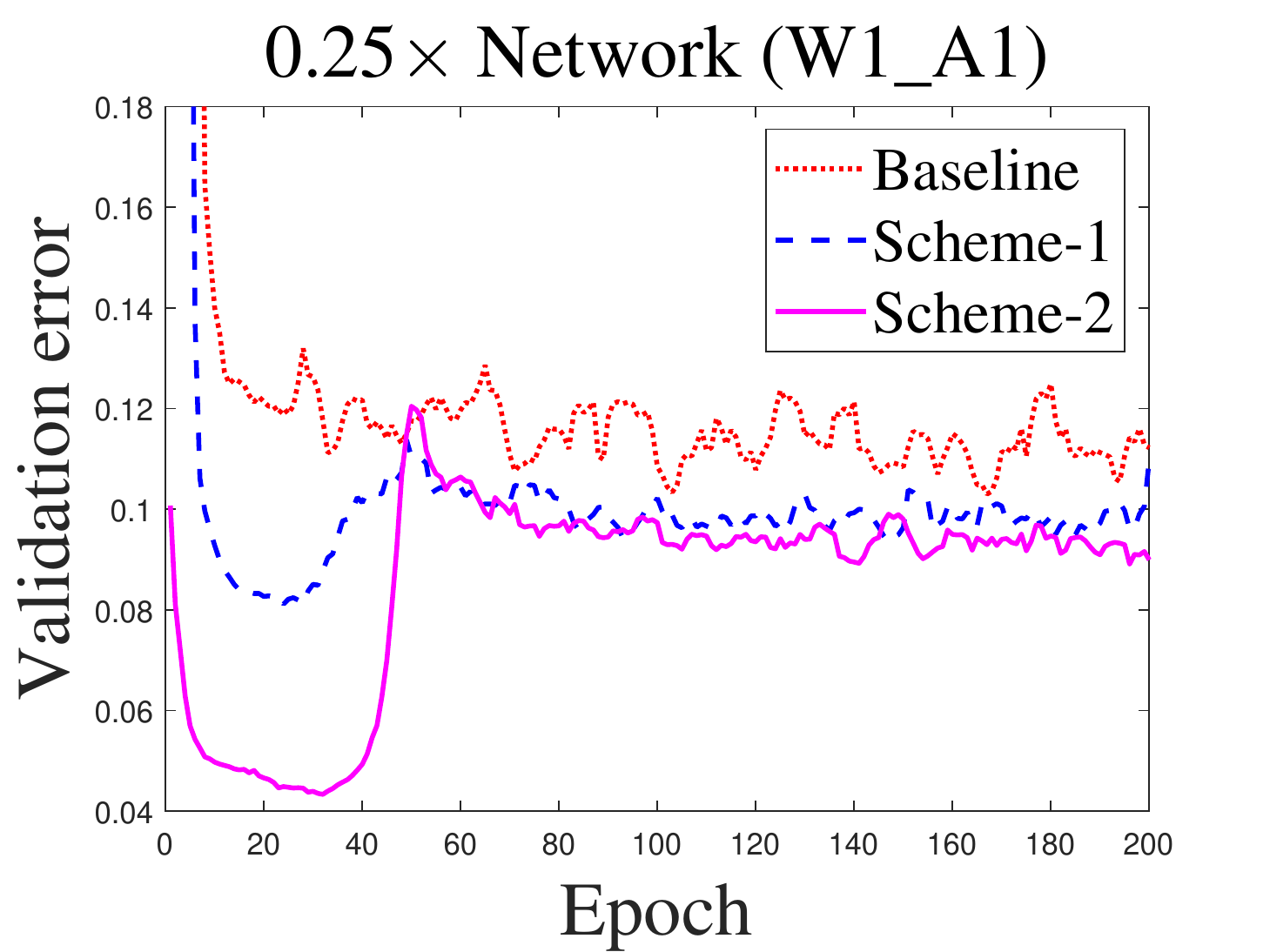}
	\end{minipage}
	\hspace{0.2cm}
	\begin{minipage}[b]{.45\linewidth}
		\centering
\includegraphics[scale =0.3]{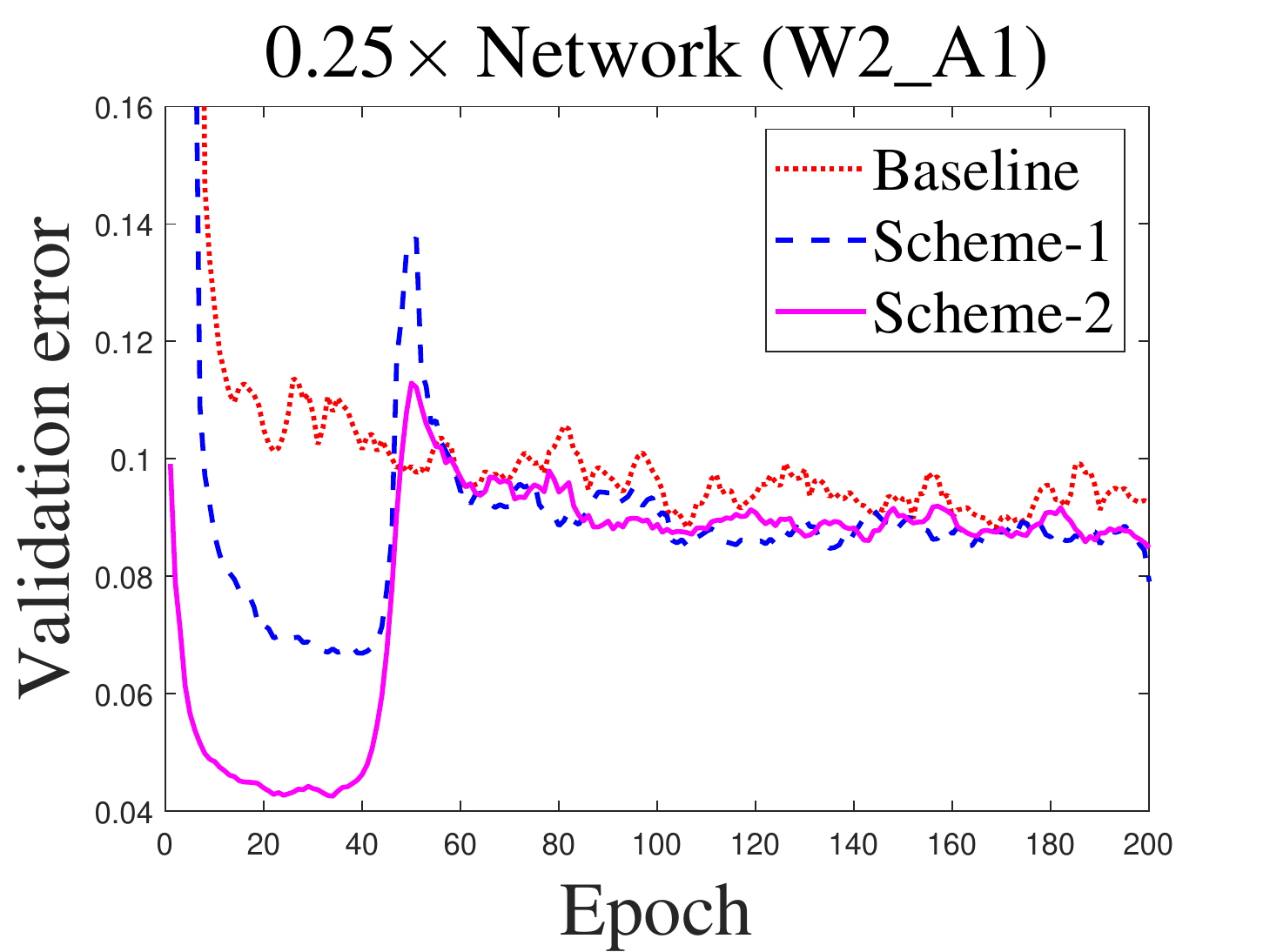}
	\end{minipage}
	\vspace{-0.1in}
	\caption{Comparison of validation errors of our two schemes based on DoReFa-Net  \cite{Zhou2016DoReFa} (left) and SYQ   \cite{faraone2018syq} (right). The  decay function is the cosine decay and decay step is set to 50 epochs.}
	\label{my_fp}
\end{figure}

On CIFAR datasets, we evaluate our two schemes for expanding low-precision networks. As shown in Table \ref{our_cifar}, our method is constantly better than the baseline method and scheme-2 is better than scheme-1. From Table \ref{repre}, we learn that the representational capability of EXP-Net using scheme-2 is more powerful than scheme-1 and the former could be regarded as a full-precision counterpart. Therefore,  the effect of scheme-2 is more significant than scheme-1.  As shown in Figure \ref{my_fp}, the EXP-Net with scheme-2 converges faster than that with scheme-1.
Another observation is that the accuracy improvement  between our schemes and the baseline method on CIFAR100 is larger than that on CIFAR10 (\emph{e.g.,} $3.11\%$ vs. $0.67\%$).
This is due to the different complexity of the two datasets,
\emph{i.e.,} the former dataset is more challenging for classification which leaves more space for improvement.

\begin{table}
	\begin{center}
		\small
		\setlength{\tabcolsep}{8pt}
		\begin{tabular}{cccccc}
			\toprule[2pt]
			\multicolumn{1}{c}{\multirow{2}{*}{Dataset}} & \multicolumn{1}{c}{\multirow{2}{*}{Method}} & \multicolumn{3}{c}{\multirow{1}{*}{Model size}} \\
			& &  0.25$\times$ & 0.5$\times$ & 1.0$\times$ \\
			\midrule[1pt]
			\multicolumn{1}{c}{\multirow{3}{*}{CIFAR10}}  &  Baseline\cite{Zhou2016DoReFa}  & 76.52 & 85.31  & 89.52 \\
			
			&  Scheme-1 &76.56 &85.68 &89.57 \\
			
			&  Scheme-2 & \textbf{77.21} & \textbf{85.85} & \textbf{90.19}  \\
			\midrule[1pt]
			\multicolumn{1}{c}{\multirow{3}{*}{CIFAR100}}  &  Baseline\cite{Zhou2016DoReFa}  & 35.68 & 51.76 & 61.43 \\
			
			&  Scheme-1 &36.81 &52.13 &62.74 \\
			
			&  Scheme-2 &\textbf{39.42}  & \textbf{55.44} & \textbf{64.54}  \\
			\bottomrule[2pt]
		\end{tabular}
	\end{center}
	\vspace{-0.2in}
	\caption{Validation accuracies (\%) for VGG-16 networks of three different sizes  with the baseline method (DoReFa-Net \cite{Zhou2016DoReFa}) and our two schemes on CIFAR10 and CIFAR100 datasets. The bitwidths for quantizing weights and activations are set to 1.}
	\label{our_cifar}
\end{table}

\subsubsection{Model Size.}

As mentioned in Table \ref{size_rp}, larger model has more powerful representational capability than a smaller model and is able to learn representations even with 1-bit weights and activations. As a result, our method could improve smaller low-precision networks more significant than the larger one. From Table \ref{svhn_size}, we could find that the gap of performance we improved becomes more and more invisible as the model size increases.  Specifically, our method can improve the baseline accuracy of 0.125$\times$ network by 1.31\% to 1.96\% while merely raises the performance of 1.0$\times$ network by 0.01\% to 0.20\%.
In addition, the performance of 0.5$\times$ and 1$\times$ low-precision networks is extremely close to that of full-precision one. Thus, it is almost impossible to improve the larger model obviously whatever quantization method or training method we use.
In addition, our method could improve the accuracy of low-precision networks under various bitwidths

\begin{table}
	\begin{center}
		\small
		\begin{tabular}{ccccccc}
			\toprule[2pt]
			\multicolumn{1}{c}{\multirow{3}{*}{\tabincell{c}{Bitwidth \\ (W/A)}}} & \multicolumn{1}{c}{\multirow{3}{*}{Method}} & \multicolumn{4}{c}{\multirow{1}{*}{Model size}} \\
			& & 0.125$\times$ & 0.25$\times$ & 0.5$\times$ & 1.0$\times$ \\
			& & (95.0) & (97.2) & (97.5) & (97.5) \\
			\midrule[1pt]
			\multicolumn{1}{c}{\multirow{3}{*}{2/1}}  &  SYQ \cite{faraone2018syq}  & 84.43 & 92.83  & 96.24 & 97.16 \\
			&  Ours & 85.74 & 93.88 & 96.54 & 97.29  \\
			& $\Delta$ & +1.31 & +1.05 & +0.30 & +0.13 \\
			\midrule[1pt]
			\multicolumn{1}{c}{\multirow{3}{*}{2/2}}  &  SYQ \cite{faraone2018syq}  & 88.39 & 94.56 & 97.04 & 97.62 \\
			&  Ours & 90.11  & 95.67 & 97.33 & 97.63   \\
			& $\Delta$ & +1.72 & +1.11 & +0.29 & +0.01   \\
			\midrule[1pt]
			\multicolumn{1}{c}{\multirow{3}{*}{2/4}}  &  SYQ \cite{faraone2018syq}  & 90.57 & 95.29 & 97.13 & 97.34 \\
			&  Ours & 92.53  & 96.46 & 97.38 & 97.54   \\
			& $\Delta$ & +1.96 & +1.17 & +0.25 & +0.20   \\
			\bottomrule[2pt]
		\end{tabular}
	\end{center}
	\vspace{-0.2in}
	\caption{Validation accuracies (\%) for four networks of different sizes  with the baseline method (SYQ \cite{faraone2018syq}) and our method on SVHN dataset. The ``W/A" values are the bits for quantizing weights/activations. ``$\Delta$" denotes the gap between the baseline and our method.}
	\label{svhn_size}
\end{table}

\subsubsection{Synchronization vs. Asynchronization.}

\begin{figure}[htbp]
	\begin{minipage}[b]{.45\linewidth}
		\centering
\includegraphics[scale =0.3]{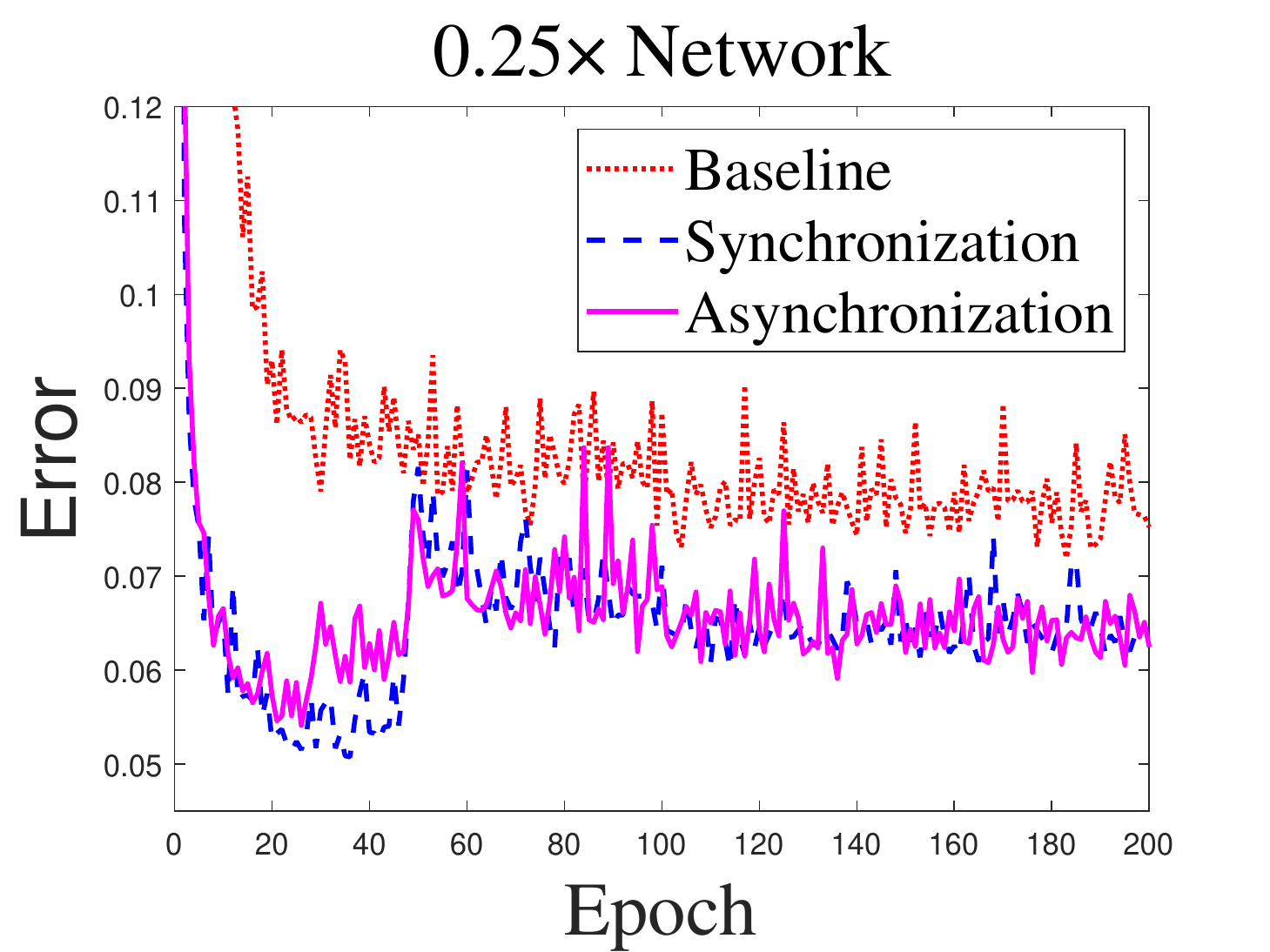}
	\end{minipage}
	\hspace{0.2cm}
	\begin{minipage}[b]{.45\linewidth}
		\centering
\includegraphics[scale =0.3]{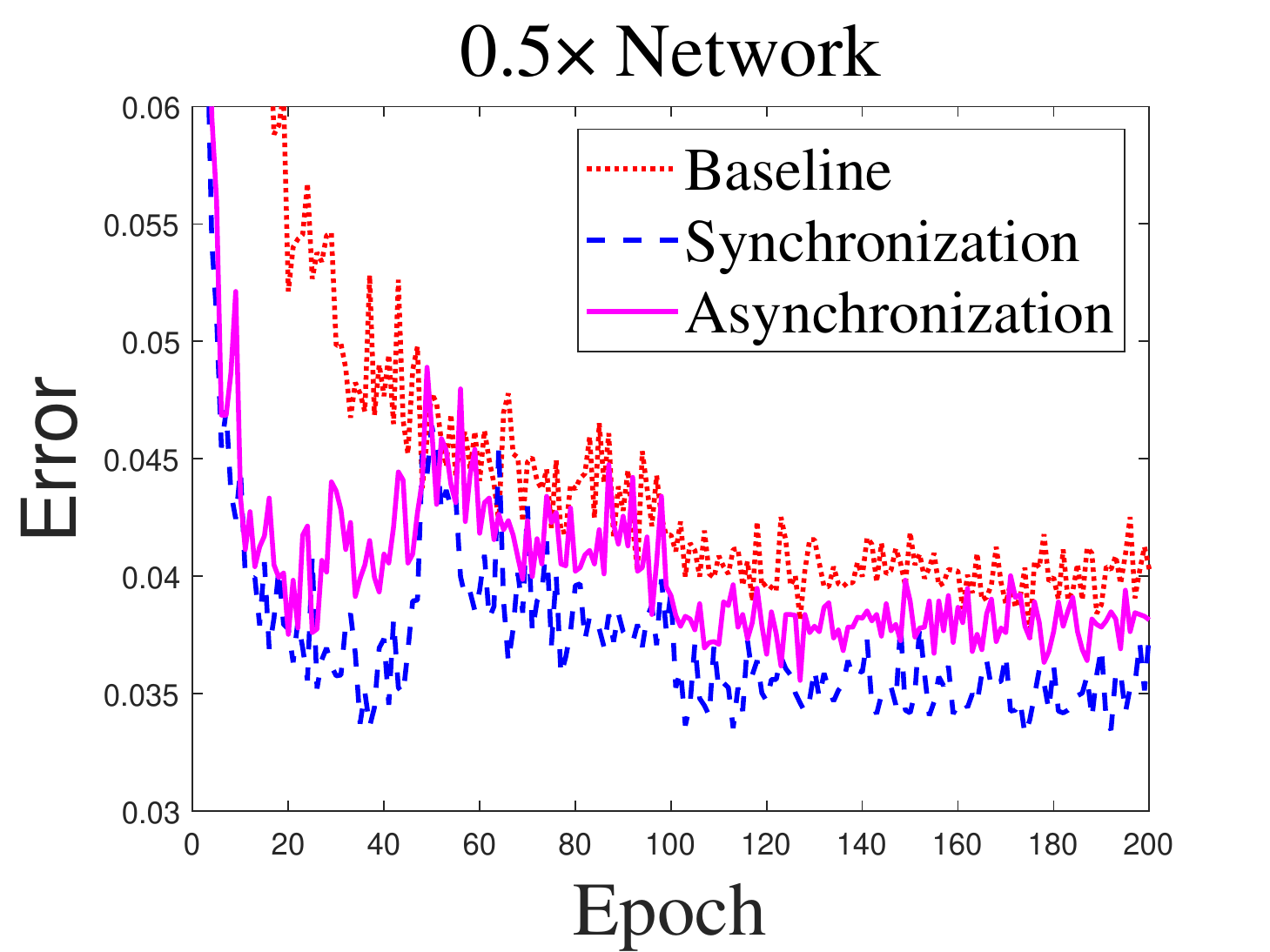}
	\end{minipage}
	\vspace{-0.1in}
	\caption{Validation errors of two different sizes of the network on SVHN dataset with the baseline method (SYQ \cite{faraone2018syq}) and our EXP-Net using synchronization and asynchronization. Weights and activations are quantized to 2-bit and 1-bit, respectively. The decay step in synchronization is set to 50 epochs while the decay steps in last three layers are set to 30, 50 and 80 epochs in asynchronization. }
	\label{svhn_synchronization}
\end{figure}

In all the above experiments, the decay functions for all full-precision layers are synchronized, \emph{i.e.,} the curves of the functions are the same. As shown in Figure \ref{svhn_synchronization}, we explore the effects of training our EXP-Net of synchronization and asynchronization.
While the full-precision output is weakened close to zero, the validation error is increasing by a visible margin. Thus, there has only once a drastic change in synchronization while there are several changes in asynchronization.
Many changes may cause our EXP-Net to  worse local minima, which is shown in Figure \ref{svhn_synchronization} (right).
Consequently, it is better to choose synchronization than asynchronization in the rest of the experiments.

\subsubsection{Expanded Structure.}

\begin{table}[t]
	\small
	\begin{center}
		\begin{tabular}{ccccccc}
			\toprule[2pt]
			Net & None & All & 123 & 456 & 56 & 6 \\
			\midrule[1pt]
			0.25$\times$ & 92.83 & 93.29 & 93.38 & 93.98 & 93.62 & 92.77  \\
			0.50$\times$ & 96.24 & 96.53 & 96.41 & 96.68 & 96.74  & 96.09   \\
			
			\bottomrule[2pt]
		\end{tabular}
	\end{center}
	\vspace{-0.2in}
	\caption{Comparison accuracies of EXP-Net using the different number of layers to expand. ``None" and ``All" denote the baseline method (SYQ \cite{faraone2018syq}) and our EXP-Net of equipping every LP layer with one FP layer, respectively. The number like ``56" in the first row means we only equip FP layers in the $5^{th}$ and $6^{th}$ LP layers. Here, we use 0.25$\times$ and 0.50$\times$ model on SVHN dataset and quantize weights/activations using 2/1-bit.}\label{svhn_layer}
\end{table}

In the above experiments, we construct our EXP-Net through equipping every low-precision layer with a full-precision layer. But expanding all LP layers is not necessary and may not the best scheme. Table \ref{svhn_layer} lists the results of expanding the different number of LP layers. It is notable that equipping every LP layer with FP one is not the best solution.
Expanding several deeper LP layers may get better performance, such as ``456" and ``56". We consider that shallow layers have more significant influence on network than deep layers. Consequently, the degradation of accuracy is significant when weakening the outputs of shallow full-precision layers gradually.
Expanding too few LP layers (\emph{e.g.,} ``6") is also not good. In this way, the improvement of the representational capability is far limited and the EXP-Net has invisible helpful to its low-precision one which results in a similar performance between our method and the baseline.

\subsubsection{Quantization Method.}

\begin{figure}[htbp]
	\begin{minipage}[b]{.45\linewidth}
		\centering
\includegraphics[scale =0.3]{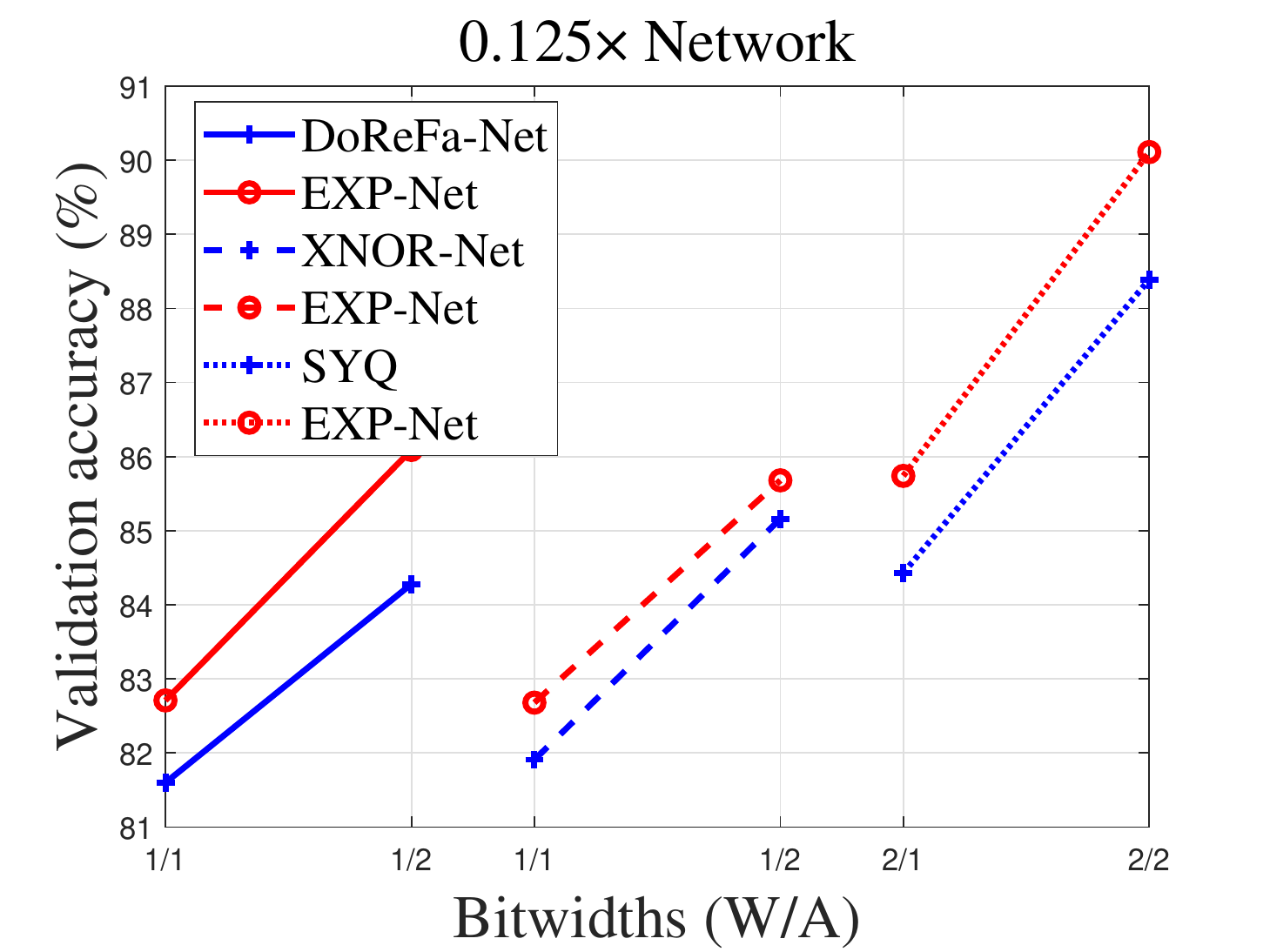}
	\end{minipage}
	\hspace{0.2cm}
	\begin{minipage}[b]{.45\linewidth}
		\centering
\includegraphics[scale =0.3]{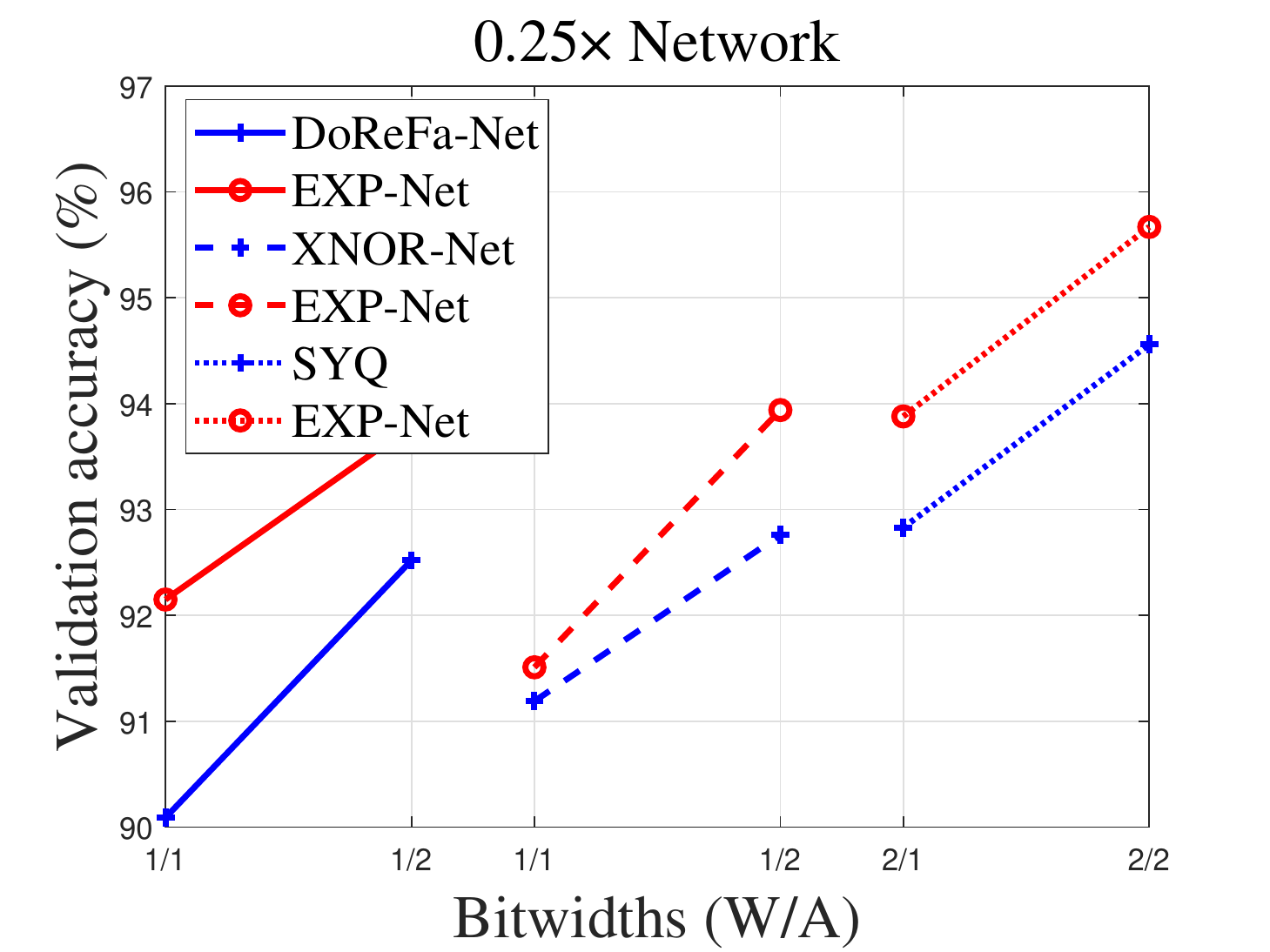}
	\end{minipage}
	\begin{minipage}[b]{.45\linewidth}
		\centering
\includegraphics[scale =0.3]{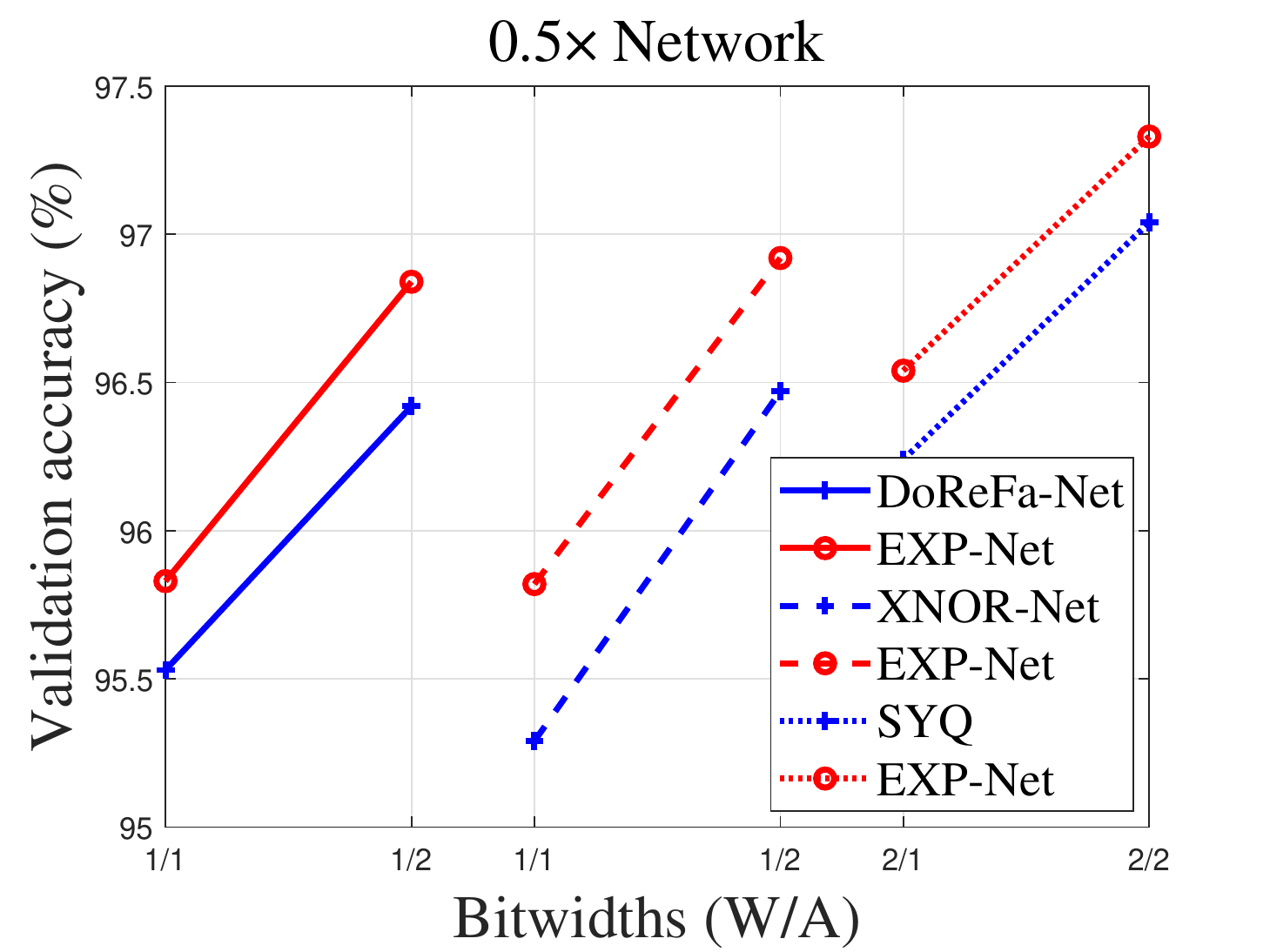}
	\end{minipage}
	\hspace{0.2cm}
	\begin{minipage}[b]{.45\linewidth}
		\centering
\includegraphics[scale =0.3]{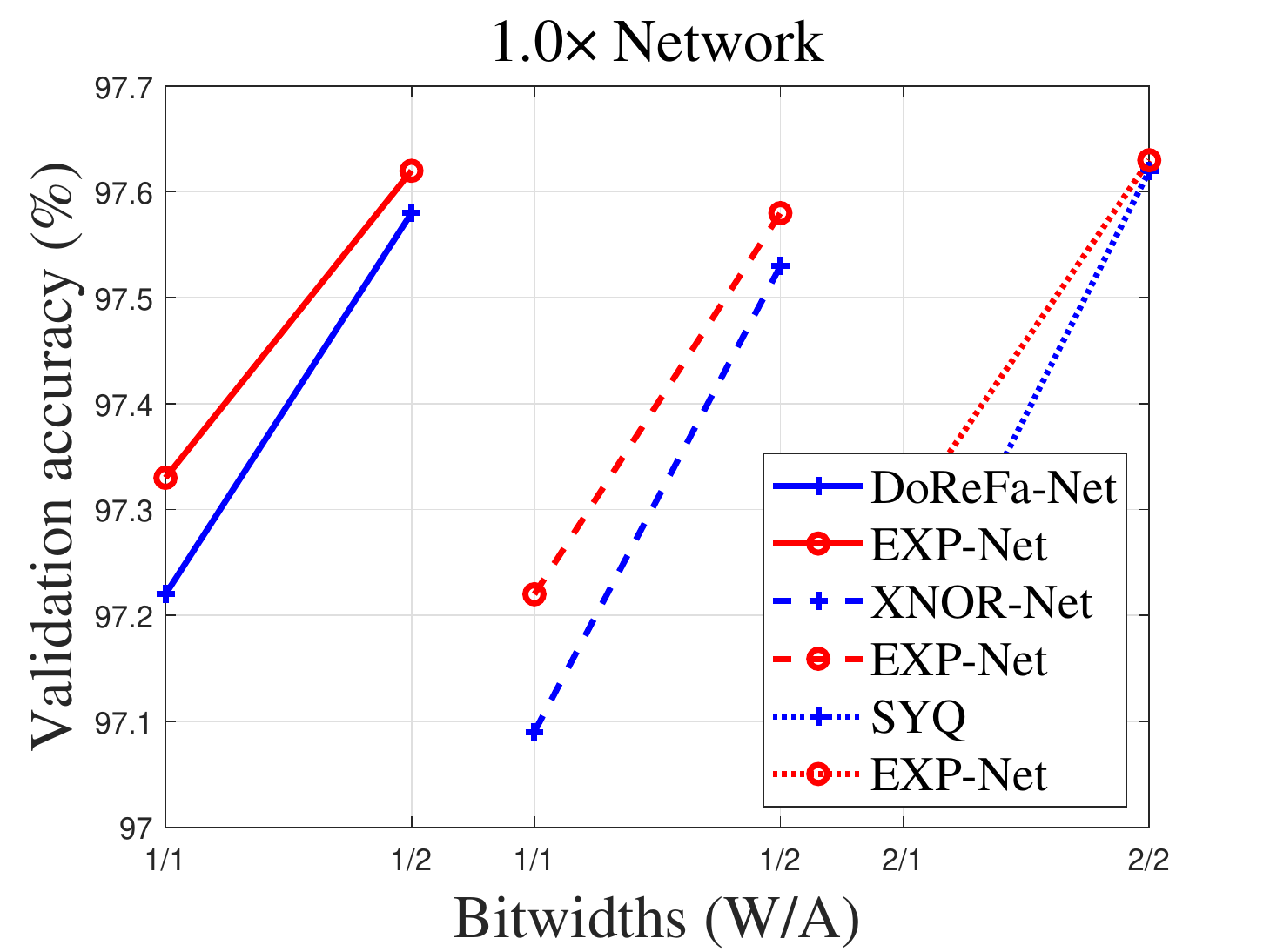}
	\end{minipage}
	\vspace{-0.1in}
	\caption{Validation accuracies of four different sizes of the network on SVHN dataset with our EXP-Net and three other low-precision networks including DoReFa-Net \cite{Zhou2016DoReFa}, XNOR-Net \cite{Rastegari2016XNOR} and SYQ \cite{faraone2018syq}. The x-axis denotes the bitwidths for quantizing weights/activations. Our method is consistently better under all settings. }
	\label{quantization_method}
\end{figure}

Since our method is a training scheme, the quantization function in our EXP-Net could be any functions, such as DoReFa-Net \cite{Zhou2016DoReFa}, XNOR-Net \cite{Rastegari2016XNOR} and SYQ \cite{faraone2018syq}. As shown in Figure \ref{quantization_method}, we compare the proposed EXP-Net with three low-precision networks with different bitwidths and network sizes.
We have two major observations from Figure \ref{quantization_method}.
First, a better quantization function can improve the performance of low-precision networks by a large margin. Specifically, on the 0.25$\times$ network, SYQ outperforms the accuracy of DoReFa-Net by 2.74\%.
Second, our method is consistently better than the standard training method with three quantization functions.

\subsection{Evaluation on ILSVRC-2012}

After analyzing different ways of training our EXP-Net,
We  evaluate our proposed method on AlexNet  for large scale image classification. Quantization functions used in EXP-Net are DoReFa \cite{Zhou2016DoReFa}, XNOR \cite{Rastegari2016XNOR}.
From Table \ref{exp_alexnet}, first, we can find that our method is consistently better than the baseline method which trains the standard low-precision networks. Specifically, our EXP-Net based on DoReFa-Net reaches a top-1 accuracy of 44.83\% which outperforms DoReFa-Net by 1.23\%.
Second, with DoReFa or XNOR quantization method, our EXP-Net cannot perform better than HWGQ\cite{cai2017deep} and SYQ\cite{faraone2018syq}. We consider that the quantization method we used is not better than that in the latter, we will integrate better quantization functions to improve our network performance.

\begin{table}[htbp]
	\small
	\begin{center}
		\setlength{\tabcolsep}{10pt}
		\begin{tabular}{cccc}
			\toprule[2pt]
			\multicolumn{1}{c}{\multirow{2}{*}{Method}} & \multicolumn{1}{c}{\multirow{2}{*}{\tabincell{c}{Bitwidth\\(W/A)}}} & \multicolumn{2}{c}{\multirow{1}{*}{Accuracy (\%)}}  \\
			& & Top-1 & Top-5 \\
			\midrule[1pt]
			Full-Precision \cite{Krizhevsky2012ImageNet} & 32/32 & 57.10 & 80.20   \\
			BNN \cite{Courbariaux2016Binarized} & 1/1 & 41.80 &  67.10   \\
			XNOR-Net$^\S$ \cite{Rastegari2016XNOR} & 1/1 & 44.20 & 69.20  \\
			DoReFa-Net$^\S$ \cite{Zhou2016DoReFa} & 1/1 & 43.60 & - \\
			DoReFa-Net$^\S$ \cite{Zhou2016DoReFa} & 1/2 & 49.80 & - \\
			DoReFa-Net$^\S$ \cite{Zhou2016DoReFa} & 1/4 & 53.00 & - \\
			QNN \cite{hubara2017quantized} & 1/2 & 51.00 & 73.70  \\
			HWGQ \cite{cai2017deep} & 1/2 & 52.7 & 76.3 \\
			SYQ$^\S$ \cite{faraone2018syq} & 1/2 & 55.40 & 78.60 \\
			\midrule[1pt]
			EXP-Net (DoReFa) & 1/1  & 44.83 & 69.73 \\
			EXP-Net (DoReFa) & 1/2  & 51.97 & 75.94 \\
			EXP-Net (XNOR) & 1/1    & 44.79 & 69.71 \\
			
			\bottomrule[2pt]
		\end{tabular}
	\end{center}
	\caption{Comparison with state-of-the-art low-precision networks on AlexNet. The ``W/A" values are the bits for quantizing weights/activations. ``EXP-Net (DoReFa)" denotes our EXP-Net used the quantization function proposed in DoReFa-Net. $^\S$ Results are trained with the pre-trained FP model initialized.}\label{exp_alexnet}
\end{table}

\section{Conclusion}

In this paper, we have presented a new learning strategy to progressively train low-precision networks. This is achieved by expanding a low-precision network during training and removing the expanded parts  for network inference. We propose two schemes to construct the expanded network structure (\emph{i.e.,} EXP-Net) by combining the low-precision and full-precision convolutional outputs and a decay method to reduce the added full-precision outputs gradually. With the full-precision parts, the representational capability of EXP-Net is significantly improved which can implicitly guide the learning of low-precision network. Extensive experimental results demonstrate the  effectiveness of our method in learning low-precision networks.

{\small
\bibliographystyle{ieee}
\bibliography{egbib}
}

\end{document}